\documentclass[a4paper,twoside]{article}
\usepackage{graphicx} 
\usepackage{booktabs}
\usepackage{times}
\usepackage{latexsym}
\usepackage{subcaption}
\usepackage{enumitem}
\usepackage{epsfig}
\usepackage{subcaption}
\usepackage{calc}
\usepackage{amssymb}
\usepackage{amstext}
\usepackage{amsmath}
\usepackage{amsthm}
\usepackage{multicol}
\usepackage{pslatex}
\usepackage{natbib}
\usepackage{algorithm2e}
\usepackage[bottom]{footmisc}  
\usepackage[T1]{fontenc}

\usepackage{placeins}
\usepackage[utf8]{inputenc}
\usepackage{multirow}
\usepackage{microtype}
\usepackage{algorithm2e}
\usepackage[bottom]{footmisc}
\usepackage{inconsolata}

\usepackage{graphicx}
\usepackage[acronym]{glossaries}
\usepackage{SCITEPRESS}

\makenoidxglossaries
\newacronym{ml}{ML}{machine learning}
\newacronym{zs}{ZS}{zero-shot}
\newacronym{fs}{FS}{few-shot}
\newacronym{os}{OS}{one-shot}
\newacronym{aa}{AA}{Authorship Attribution}
\newacronym{av}{AV}{Authorship Verification}
\newacronym{llm}{LLM}{Large Language Model}
\newacronym{nlp}{NLP}{natural language processing}

\begin{document}

\title{Author Representation Strategies for Zero-Shot Authorship Attribution: A Comparative Study of LLM-Based and Embedding-Based Approaches}

\author{
\authorname{
Nudrat Habib\orcidAuthor{0009-0005-9310-1701},
Tosin Adewumi\orcidAuthor{0000-0002-5582-2031},
Sana Sabah Al-Azzawi\orcidAuthor{0000-0001-7924-4953},
Marcus Liwicki\orcidAuthor{0000-0003-4029-6574} and
Elisa Barney\orcidAuthor{0000-0003-2039-3844}
}
\affiliation{Luleå University of Technology, Sweden}
\email{firstname.lastname@ltu.se}
}


\abstract{\acrfull{aa} requires capturing fine-grained stylistic characteristics, making it particularly challenging in \acrfull{zs} settings where no task-specific supervision is available. In this work, we investigate the effect of author representations on ZS AA by evaluating a label-only prompting baseline together with three author representation strategies: representative writing samples, LLM-generated descriptions, and style embeddings (LISA). The first three approaches perform attribution using LLM prompting, while the embedding-based approach uses style embeddings with cosine similarity.
We investigate the influence of prompt design and propose a two-stage embedding-based attribution framework that combines candidate-space reduction with embedding-dimension selection. The results show that label-only \acrshort{zs} \acrshort{aa} is ineffective, while incorporating author-specific representations consistently improves attribution performance. Among the evaluated approaches, the proposed two-stage LISA framework achieves the strongest overall performance, whereas LLM-generated style descriptions provide a substantially more compact representation of author style at the cost of some attribution performance. These findings demonstrate the importance of author representation in \acrshort{zs} \acrshort{aa}, while indicating that current open-source \acrshortpl{llm} remain insufficient for robust attribution without more effective representation learning.}

\onecolumn \maketitle \normalsize \setcounter{footnote}{0} \vfill

\section{Introduction}
\acrfull{aa} is the process of determining the likely author of a given text by analyzing linguistic patterns, stylistic cues, and structural characteristics \cite{huang2025authorship}. It has important applications in areas such as forensic linguistics, cybersecurity (e.g., fraud and cybercrime detection), plagiarism detection, and digital content authentication \cite{habib2025trends}. Conventional \acrshort{aa} methods are predominantly supervised, requiring labeled examples for each candidate author. Although effective in controlled environments, such approaches are data-intensive and often struggle to generalize across domains, genres, or writing conditions. This limitation motivates the exploration of alternative paradigms that reduce reliance on annotated data. Several paradigms have been proposed to reduce this dependency. In \acrfull{fs} learning, the model is given a small number  of task demonstrations K, while \acrfull{os} represents the special case where K=1. In contrast, \acrfull{zs} relies solely on a natural language description of the task without any labeled examples (K=0) \cite{brown2020language}. 

Recent advances in \acrshortpl{llm} have made \acrshort{zs} inference practical across a wide range of \acrshort{nlp} tasks, making them a promising direction for \acrshort{aa} in low-resource or zero-supervision settings. In the \acrshort{zs} setting, \acrshortpl{llm} leverage knowledge acquired during pre-training together with instruction tuning, alignment mechanisms such as Reinforcement Learning from Human Feedback (RLHF), and the task prompt to perform inference. Together these mechanisms enable models to follow task instructions, generalize to unseen tasks, and perform in-context reasoning, leading to strong performance across tasks including machine translation, question answering, summarization, and sentiment analysis \cite{radford2019language}.

Despite strong performance in \acrshort{zs} settings on tasks aligned with pretraining knowledge \cite{brown2020language}, the applicability of LLMs in ZS setting to tasks requiring fine-grained stylistic discrimination remains unclear. The LLMs are sensitive to prompt design and may struggle without additional guidance. Several limitations of \acrshort{zs} learning across \acrshort{nlp} tasks have been identified, particularly in settings that require complex reasoning, fine-grained evaluation, and robustness to domain shifts, where models are sensitive to prompt formulation, prone to hallucination, and less reliable in domain-specific or low-resource scenarios due to gaps in pre-training coverage \cite{ramesh2025review}. Although \acrshort{aa} is not explicitly discussed, the need for stable and subtle stylistic discrimination in \acrshort{aa} aligns with the types of tasks identified as challenging for \acrshort{zs} approaches. For such challenging tasks, variants of \acrshort{zs} prompting have been proposed. One such formulation distinguishes between \acrshort{zs} settings with and without definitions in the context of fallacy classification \cite{pan2024llms}. In the definition-free setting, the model receives only label names and must infer their meanings implicitly. In the definition-based setting, each label includes a natural-language explanation, providing semantic guidance without using labeled examples, thus preserving the \acrshort{zs} nature of inference.

Following the formulation in \cite{pan2024llms}, we distinguish between two \acrshort{zs} settings in this work. We define a true \acrshort{zs} setting, where only candidate author labels are provided, and a definition-enhanced \acrshort{zs} setting, where candidate authors are supplemented with author representations but no labeled attribution examples or task-specific fine-tuning are used. Under this definition, both settings preserve the \acrshort{zs}  nature of inference. We then investigate these settings for \acrshort{aa} through different author representation strategies. Our goal is to evaluate whether \acrshort{aa} can be performed without supervised training and whether providing explicit author representations improves attribution performance. 

\textbf{This work aims to address the following research questions.}
\begin{itemize} [noitemsep, topsep=0pt]

    \item \textbf{RQ1:} Can \acrshort{aa} be performed effectively in a true \acrshort{zs} setting without task-specific fine-tuning?

    \item \textbf{RQ2:} Does augmenting \acrshort{zs} \acrshort{aa} with author-related definitions improve attribution performance, and how do different definition strategies compare?
    \item \textbf{RQ3:} How do different prompt formulations influence author description generation and \acrshort{zs} \acrshort{aa} performance?
\end{itemize}
\textbf{Our contributions are as follows:}
\begin{itemize} [noitemsep, topsep=0pt]
    \item We evaluate \acrshort{aa} using LLMs in a true \acrshort{zs} setting without task-specific 
    fine-tuning or labeled training data. 
    
    \item We adapt and extend the definition-enhanced \acrshort{zs}  paradigm to \acrshort{aa} by extending traditional label-only prompting with three definition strategies: sample-based definitions, LLM-generated descriptions, and style-based embeddings.
    
    \item We evaluate the impact of different definition strategies on attribution performance, providing insights into their effectiveness for capturing author-specific stylistic signals.
    
    \item We investigate the role of prompt design in both author description generation and \acrshort{aa}. We design four prompts for generating author descriptions and evaluate them together with four attribution prompts adapted from prior work.
    
    \item We present a unified evaluation framework for comparing label-only prompting and three definition strategies under consistent experimental settings for \acrshort{zs} \acrshort{aa}.
\end{itemize}

The rest of the paper is organized as follows. \textbf{Section 2}  presents the literature review, \textbf{Section 3} describes the methodology, \textbf{Section 4} presents the results and discussion and  \textbf{Section 5} concludes the paper. 

\section{Literature Review}
\acrshort{llm}s have been widely studied in \acrshort{zs} and \acrshort{fs} settings across a wide range of  \acrshort{nlp} tasks.
\citet{chamieh2024llms} study \acrshort{llm} performance in \acrshort{zs} and \acrshort{fs} settings for short answer scoring, showing variability in \acrshort{zs} results and difficulty in capturing fine-grained distinctions reliably. 
\citet{ramesh2025review} survey \acrshort{zs} and \acrshort{fs} learning in \acrshort{nlp}, noting that while \acrshort{zs} reduces reliance on labeled data, its performance depends heavily on pretraining, prompt design, and domain conditions. The study highlights issues such as instability, hallucination, and reduced robustness, along with a gap between open- and closed-source models, suggesting that \acrshort{zs} methods often require adaptation for reliable performance. These studies suggest that, although \acrshort{llm}s show strong potential in general \acrshort{zs} settings, their reliability in tasks requiring fine-grained distinctions such as \acrshort{aa} remains limited, motivating the need for further investigation in this direction.

Recent work has explored the use of LLMs for AA under different settings.  \citet{huang2024can} explore \acrshort{aa} using \acrshort{llm}s with Linguistically Informed Prompting (LIP), where author samples serve as style definitions. Results on the Enron Email \cite{klimt2004enron} and Blog Authorship \cite{schler2006effects} datasets show improved stylistic recognition and interpretability without task-specific fine-tuning. In this setting, they report weighted F1 scores for traditional and neural methods in the range of 11–30\%, depending on the choice of model (TF-IDF, BERT). Using LIP, the authors report a weighted F1 score of 30.56\% with GPT-3.5 and 84.45\% with GPT-4. In contrast, the reported performance drops to 10.56\% when using the open-source Mistral model. \citet{hu2024bayesian} perform  \acrshort{os} \acrshort{aa} using a Bayesian method over LLM probability outputs. They evaluate models such as LLaMA 3 and GPT, achieving 85\% accuracy on the IMDB62 and Blog datasets, and use \cite{huang2024can} as a baseline. Open-set \acrshort{aa} is studied by \citet{tan2025open} using a two-stage method consisting of a candidate selection stage to retrieve related documents and then  the top 20 candidates are selected for the final attribution decision in a second stage.  While these studies demonstrate the potential of \acrshort{llm}s for \acrshort{aa}, they often rely on structured prompting, probabilistic modeling, or multi-stage pipelines, making it difficult to isolate the effectiveness of LLMs in a true \acrshort{zs} setting.

Building on this, recent work has highlighted important limitations of LLMs for \acrshort{aa}. 
\citet{gorovaia2024sui} investigate the use of \acrshort{llm}s for \acrshort{aa} in historical Latin texts, evaluating their ability to capture stylistic and semantic cues. The study highlights limitations in modeling subtle stylistic signals, especially when semantic content dominates. \citet{adewumi2025limitations} show that \acrshort{zs} \acrshort{aa} with open-source LLMs is unstable, with frequent misattributions and fabrications. This suggests potential reliability concerns that may extend to more challenging attribution scenarios.  
Collectively, these findings highlight that \acrshort{llm}s face inherent challenges 
motivating a more controlled investigation of \acrshort{zs} \acrshort{aa} with different 
strategies.

Prompt-based approaches have been explored to guide LLMs in tasks requiring structured reasoning or stylistic comparison. In particular, definition-based prompting provides additional semantic grounding by supplying label descriptions or example-based definitions. \citet{pan2024llms} explore \acrshort{zs} classification with and without definitions, showing that providing natural language descriptions alongside labels can guide model behavior without requiring training data. The LIP approach proposed by \citet{huang2024can} similarly incorporates author samples as definitions. These methods demonstrate that structured contextual information can influence model behavior, although their effectiveness for \acrshort{zs} AA remains limited motivating further investigation. 

Embedding-based approaches have been widely explored for stylistic representation in \acrshort{aa}. \citet{benzebouchi2019authors} use Word2Vec-based author and document representations with neural classifiers, highlighting early embedding-based methods for capturing stylistic information. 
\citet{riley2021textsettr} propose TextSETTR, a label-free few-shot style transfer method based on T5 that learns style vectors from adjacent sentences. \citet{terreau2024capturing} introduce VADES, a Variational Information Bottleneck-based architecture that learns stylistically constrained embeddings for authors and documents. \citet{malik2024empirical} introduce LISA embeddings, where stylistic features are represented as high-dimensional vectors. They construct 768-dimensional style representations and select the most prominent stylistic features for each text. Their work analyzes how LLMs generate text under different socio-demographic personas, showing that models such as GPT-3.5, LLaMA, and Mixtral exhibit distinct stylistic variations. 
\citet{alshomary2025latent} explore latent stylistic representations for \acrshort{aa} through clustering and LLM-generated style descriptions, showing that while they capture stylistic signals, they are not inherently interpretable. While embedding-based approaches show promise in capturing stylistic information, their effectiveness in \acrshort{zs} settings remains unclear. In this work, we build on LISA embeddings for \acrshort{aa} using similarity-based attribution based on vector representations of writing style.

While prior work has explored \acrshort{zs} learning, prompting strategies, and embedding-based methods for \acrshort{aa}, a clear understanding of how these approaches interact in \acrshort{zs} settings remains limited. In particular, the role of different definition strategies and their impact on capturing fine-grained stylistic distinctions is not well understood. Furthermore, the combined effect of prompt design and representation choices has not been systematically analyzed. In this work, we address these gaps by evaluating \acrshort{zs} \acrshort{aa} across different prompting, definition-based, and representation settings.

\section{Methodology}
To evaluate \acrshort{zs} \acrshort{aa}, we investigate two experimental settings. The first, described in Section 3.1, is a label-only setting that provides the model with candidate author labels and asked to identify the author of the query text without additional guidance. The second is a definition-based setting (Sections 3.2–3.4), where each candidate author is supplemented with additional information to facilitate stylistic discrimination. We investigate three definition strategies, where the first one (Section 3.2) is based on the approach proposed in  \cite{huang2024can}  and the other two (Sections 3.3 and 3.4) are proposed in this study. 

The dataset used is the \textbf{Blog Authorship Corpus}, which consists of posts from 19,320 bloggers collected from blogger.com in August 2004 
\cite{schler2006effects}. This dataset has been widely used in the literature for \acrshort{aa} using traditional machine learning, deep learning, and, more recently, \acrshort{llm}s. Due to its large size, different studies often use different subsets, making results not directly comparable. To evaluate the sample-based definition strategy, additional subsets of the dataset were created using different sampling strategies. For inter-configuration comparison, the same subset of the dataset must be used to ensure a fair comparison. In our case, we adopt the same subset used in \cite{huang2024can}, enabling a more direct comparison of different configurations. 

The evaluation metrics include accuracy (Acc),
Weighted F1 (W-F1),  Micro-averaged F1 ($\mu$F1), Macro-averaged F1(M-F1) and confusion matrices. 
W-F1 gives an average F1 score weighted by class size. Micro F1
calculates the overall average F1 score, combining all classes, and is sensitive to class imbalance.
Macro F1 computes the unweighted average of F1
scores across classes, treating each class equally,
ideal for assessing minority class performance \cite{huang2024can}.
The label-only, sample-based definition, and description-based definition experiments were evaluated using three open-source instruction-tuned LLMs: Mixtral-8x7B-Instruct, Gemma-2-9B-IT, and Qwen2.5-7B-Instruct. All three models were evaluated under identical experimental settings to ensure a fair comparison.
In the subsections below, we discuss these different experimental configurations in detail. 

\subsection{Label-Only ZS AA}
In this setting  the model is provided only with the query text and the candidate label space, without any additional definitions or semantic guidance. In this configuration, the model must infer the correct author solely based on the prompt instructions and its pretrained knowledge. The prompt used is shown in Table \ref{tab:zs_label_only_prompt} in Appendix \ref{app:prompts}. We did analysis for different subsets including 3, 5, 10 and 15 candidate authors.

\subsection{Definition-Enhanced ZS AA: Samples as definition}
The first definition-enhanced \acrshort{zs} \acrshort{aa} strategy we evaluate uses unlabeled author samples as implicit definitions, enabling the model to compare a query text against representative examples of each candidate author. \cite{huang2024can} implemented this idea through their LIP approach, where author samples were supplied as definitions to guide the LLM in assessing stylistic similarity. 
 In this setup, one writing sample of each author was provided along with the query text 
to help the model compare the stylistic characteristics of the query text against representative writing samples from each candidate author.
 
 We evaluated the approach under three different author-selection settings while keeping the number of authors and predictions per author fixed (10 authors, 3 predictions per author). Specifically, we experimented with (i) the subset provided in the original repository, (ii) the top 10 authors by frequency,  and (iii) a random selection of 10 authors. To ensure consistency and avoid extremely short fragments, we restricted both query texts and definition samples to a minimum token length of 50 and enforced uniqueness of selected samples within each CSV file. For AA \cite{huang2024can} designed 4 prompts as shown in Table \ref{tab:aa_prompts} in Appendix \ref{app:prompts} and these prompts were used here.
\subsection{Definition-Enhanced ZS AA: Descriptions as definition}
In this setup, we replace samples with a generic description of the author’s writing style. These descriptions are generated by prompting the model with representative writing samples and asking it to summarize the author's writing style rather than the content of the text. Ideally, multiple samples per author would produce more robust style representations; however, to ensure a fair comparison across all definition strategies, we fix all other parameters and use only one sample for description generation.
\begin{itemize} [noitemsep, topsep=0pt]
    \item \textbf{Model used:} For description generation, we use Qwen2.5-7B-Instruct \cite{qwen2}, an instruction-tuned \acrshort{llm}, in a ZS setting without task-specific finetuning or examples. 

    \item \textbf{Prompts:}
    For description generation we designed 4 different prompts  shown in Table \ref{tab:description_prompts} Appendix \ref{app:prompts}. 
    
\item \textbf{AA prompt adjustment}: We adjusted the AA prompts to use descriptions instead of samples.
 
\end{itemize}

\subsection{Definition-Enhanced ZS AA: Style-based embeddings as definition} 
Finally, for definition-enhanced \acrshort{zs} \acrshort{aa} using style-based embeddings, we used the interpretable style embedding model LISA introduced in \cite{patel2023learning}. LISA is built by distilling GPT-3-generated stylometric annotations into an EncT5-based model that produces human-interpretable style representations. It generates a 768-dimensional style vector, where each dimension corresponds to a textual style attribute such as “the author uses simple language”, “the author uses a negative tone”, and “the author uses offensive language”. Each dimension takes a value in the range [0,1], representing the probability that the corresponding stylistic attribute is present in the text.  To ensure consistency with the other definition strategies, a single representative writing sample was selected for each candidate author. LISA embeddings were generated independently for both the query text and each candidate writing sample. For every query, cosine similarity was computed between the query embedding and the embeddings of all candidate authors. 
We conducted experiments in two settings: a single-stage approach and a two-stage approach. In the single-stage setting, all 768 dimensions are used directly for classification and the candidate with the highest cosine similarity to the query embedding was selected as the predicted author. In the two-stage process, all 768 dimensions are first used to rank the candidate authors based on cosine similarity. The Top-k Recall is then computed to evaluate whether the true author is retained among the top-k candidates. In the second stage, cosine similarity is recomputed using only a selected subset of the embedding dimensions among the retained top-k candidates. In this study, we explored top-$N$, bottom-$N$, and a combination of both top-$N$
 and bottom-$N$ dimensions.

 \section{Results and discussion}
In this section, we present  the results obtained across all experiments and further compare the performance of the different definition-based methods.
\subsection{Label-Only ZS AA}
We first evaluate label-only \acrshort{zs} \acrshort{aa} quantitatively across different candidate-set sizes and then analyze the prediction behavior of the evaluated \acrshort{llm}s through confusion matrices, focusing on prediction concentration and selection bias.

\subsubsection{Impact of increasing number of authors}
Label-only \acrshort{zs} \acrshort{aa} provides the model with very limited task-specific information, requiring it to attribute authorship solely from the query text and the candidate author labels using LLM's built-in pretraining information. To investigate how the size of the candidate space affects attribution performance, we evaluated three open-source instruction-tuned \acrshort{llm}s on nested subsets containing 3, 5, 10, and 15 candidate authors, with 10 query texts per author. The results, Table~\ref{tab:zs_results}, show that across all three models, attribution accuracy decreases as the number of candidate authors increases. This trend is consistent with observations in multi-class classification and \acrshort{zs} evaluation, where performance typically decreases as the label space expands \cite{habib2025trends}. Gemma exhibits the highest accuracy across all candidate-set sizes, declining from 43.3\% (3 authors) to 8.0\% (15 authors). Mixtral shows the steepest degradation, with accuracy decreasing from 30.0\% to 4.6\%. Although Qwen exhibits a slight increase between the 10- and 15-author settings (6.0\% to 7.3\%). This minor fluctuation does not alter the overall trend that larger candidate sets substantially increase the difficulty of label-only \acrshort{zs} \acrshort{aa}. Since the subsets are nested, the observed performance degradation cannot be attributed to changes in the underlying data distribution but is primarily associated with the increasing number of candidate authors. Without task-specific fine-tuning or additional stylistic guidance, the models must discriminate among a progressively larger set of stylistically similar candidates, making accurate attribution considerably more challenging.

 In this pure \acrshort{zs} setting, the candidate labels themselves convey no stylistic information. As a result, the model  assigns a query text to one of several anonymous identifiers without any explicit representation of the corresponding authors' writing styles. This places a fundamental limitation on attribution performance, since the model lacks the information required to perform direct stylistic comparison. Beyond the quantitative decline in attribution accuracy, the confusion matrices reveal important changes in prediction behavior as the candidate space expands. These observations are discussed in the following subsection.
 
\subsubsection{Prediction Concentration and Selection Bias}
Prediction bias has been widely observed in \acrshort{llm}s, where models exhibit systematic preferences toward a subset of candidate outputs rather than distributing predictions uniformly across all available labels. One well-studied form of prediction bias is positional bias, in which predictions depend on the ordering of candidate options within the prompt rather than their semantic content \cite{wang2025eliminating}. 
In this work, we analyze whether the evaluated models exhibit selection bias, characterized by concentrating predictions on only a small subset of candidate authors.

\begin{figure*}[t]
    \centering
    
    \begin{subfigure}[t]{0.43\textwidth}
        \centering
        \includegraphics[
            width=\linewidth,
            trim=0 0 0 0,
            clip
        ]{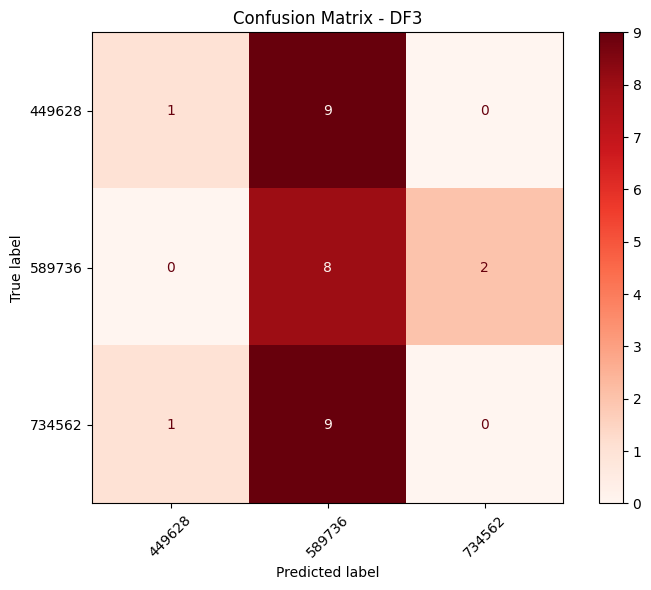}
        \caption{ 3 authors.}
        \label{fig:zs3}
    \end{subfigure}
    \hfill
    \begin{subfigure}[t]{0.43\textwidth}
        \centering
        \includegraphics[
            width=\linewidth,
            trim=0 0 0 0,
            clip
        ]{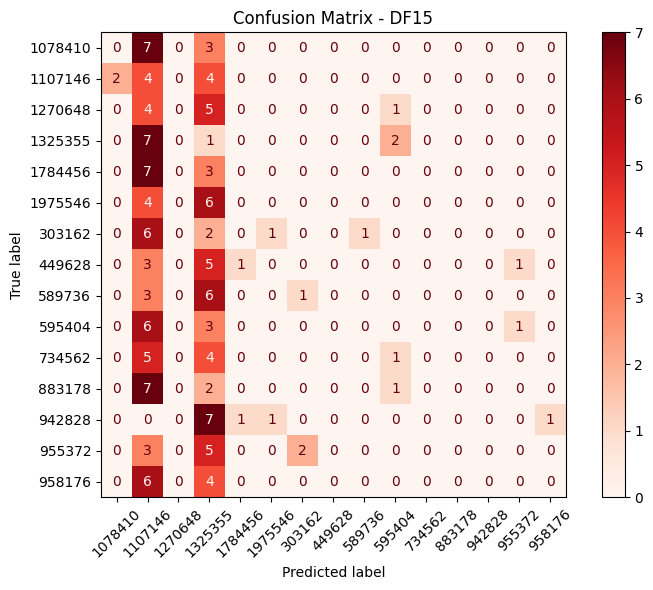}
        \caption{15 authors (stronger selection bias).}
        \label{fig:zs15}
    \end{subfigure}

    \caption{Confusion matrices for Label Only ZS \acrshort{aa} showing selection bias in Mixtral: (a) 3-authors, (b) 15-authors.}
    \label{fig:zs_combined}
\end{figure*}

\begin{figure*}[!ht]
\centering

\begin{subfigure}{0.32\textwidth}
    \centering
    \includegraphics[width=\linewidth]{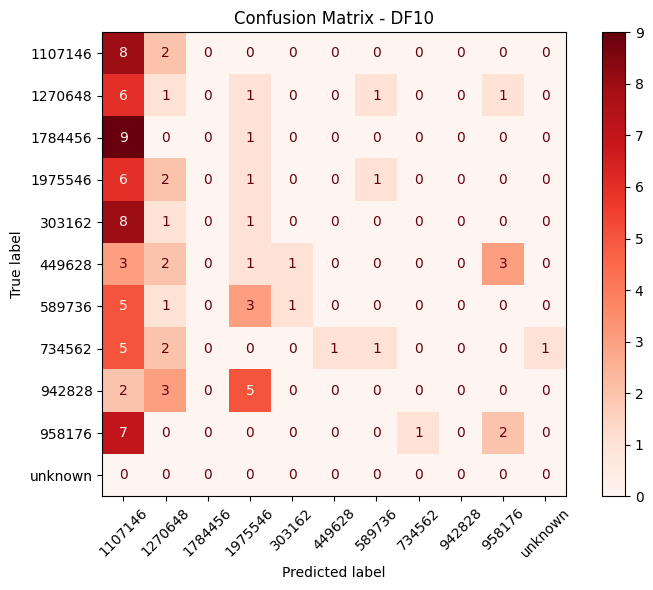}
    \caption{Mixtral-8x7B-Instruct}
\end{subfigure}
\hfill
\begin{subfigure}{0.32\textwidth}
    \centering
    \includegraphics[width=\linewidth]{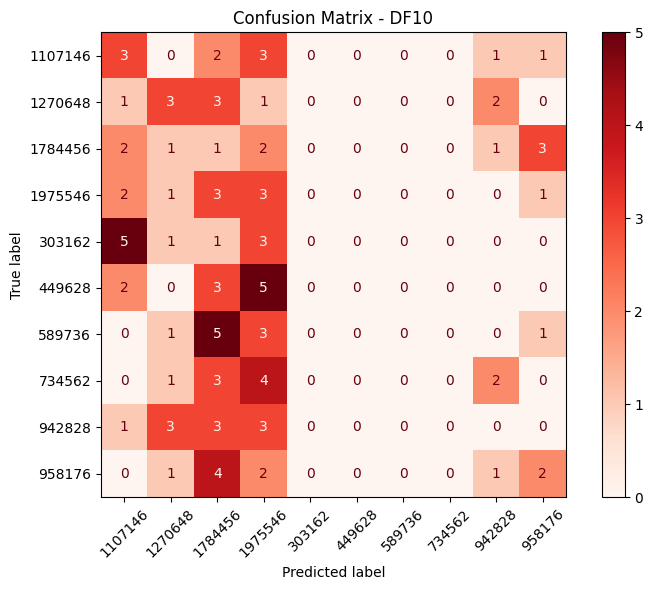}
    \caption{Gemma-2-9B-IT}
\end{subfigure}
\hfill
\begin{subfigure}{0.32\textwidth}
    \centering
    \includegraphics[width=\linewidth]{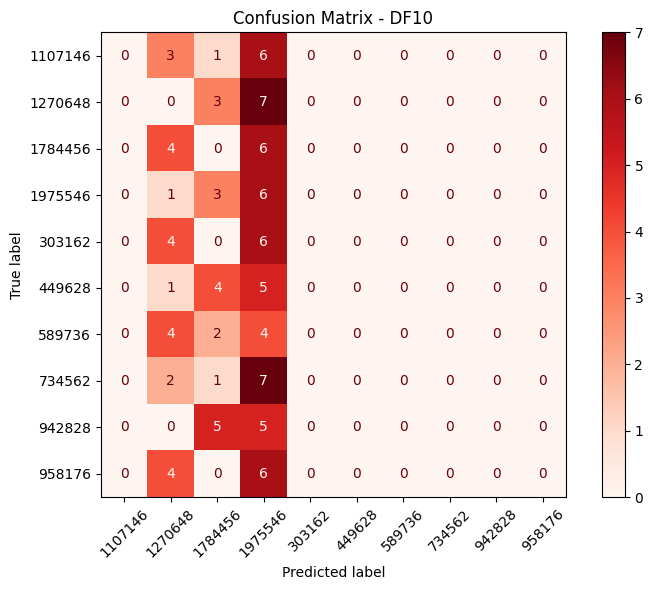}
    \caption{Qwen2.5-7B-Instruct}
\end{subfigure}

\caption{Confusion matrices for label-only \acrshort{zs} \acrshort{aa} on the 10-author subset using three open-source instruction-tuned 
\acrshortpl{llm}. All three models exhibit prediction concentration, although the distribution of predictions across candidate authors differs substantially between models.}
\label{fig:zs10_model_comparison}
\end{figure*}

Figure \ref{fig:zs3} and \ref{fig:zs15} compare the prediction behavior of Mixtral for the 3- and 15-author settings, respectively. In the 3-author setting, predictions remain relatively distributed across the candidate authors, although some concentration is already visible. As the number of candidate authors increases to fifteen, prediction concentration becomes substantially stronger, with most predictions assigned to only a small subset of candidate authors while several authors receive few or no predictions. This increasing concentration suggests that the model increasingly concentrates its predictions on a smaller subset of candidate authors as the attribution task becomes more challenging, consistent with stronger selection bias. These observations are consistent with previous studies reporting biased prediction behavior in \acrshort{llm}s, including position-sensitive behavior in multiple-choice tasks \cite{zheng2023large,
wang2025eliminating}.
Figure~\ref{fig:zs10_model_comparison} compares 
the prediction behavior of the  three evaluated \acrshort{llm}s on the same 10-author subset. All three models exhibit prediction concentration, 
however, the nature of this concentration differs across models. Mixtral exhibits a pronounced preference toward a single dominant author while still assigning occasional predictions to several other candidate authors. In contrast, Gemma distributes its predictions across a small group of dominant authors, leaving many candidate authors rarely or never predicted. Qwen shows the strongest restriction of the prediction space, with predictions largely confined to only a few candidate authors and most remaining labels receiving no predictions. These observations indicate that, although prediction concentration is present in all evaluated models, it manifests differently, affecting both the diversity and distribution of predicted author labels.

\begin{table}[t]
\centering
\caption{Label-only \acrshort{zs} \acrshort{aa} Accuracy across three open-source \acrshortpl{llm} under different numbers of candidate authors.}
\label{tab:zs_results}
\small
\setlength{\tabcolsep}{4pt}
\renewcommand{\arraystretch}{1.05}

\begin{tabular}{lccccc}
\toprule
Subset & $n_{\text{rows}}$ & $n_{\text{authors}}$ &
\textbf{Mixtral} &
\textbf{Gemma} &
\textbf{Qwen} \\
\midrule
df3  & 30  & 3  & 30.0 & \textbf{43.3} & 30.0 \\
df5  & 50  & 5  & 22.0 & \textbf{28.0} & 16.0 \\
df10 & 100 & 10 & \textbf{12.0} & \textbf{12.0} & 6.0 \\
df15 & 150 & 15 & 3.3  & \textbf{8.0} & 7.3 \\
\bottomrule
\end{tabular}
\end{table}

Taken together, these findings suggest that label-only \acrshort{zs} \acrshort{aa} provides insufficient author-specific information for reliable stylistic discrimination, particularly as the candidate space grows. This motivates expanding beyond pure ZS to explore  definition-enhanced prompting strategies in the following sections.

\subsection{Definition-Enhanced ZS AA: Samples as definition}

ZS in the purest sense implies no additional information is provided beyond the query as discussed in Section 4.1.  LLMs even with their pre-training information do not have enough capacity to successfully do AA under these conditions. We therefore investigate whether staying ZS by not training the LLM but augmenting the prompt with explicit author representations can improve attribution performance. The second set of experiments, Table~\ref{tab:zs_samples_csv_variants},  evaluates sample-based definitions under three author-selection settings: the subset provided by \cite{huang2024can}, the top 10 authors by frequency, and a random selection of 10 authors. 
Compared with the label-only \acrshort{zs} setting, the inclusion of representative author samples generally results in higher attribution performance across the evaluated \acrshort{llm}s. Although the sample-based experiments were conducted on different author subsets and therefore are not directly comparable to the label-only setting, the overall trend suggests that providing representative writing samples supplies useful stylistic context beyond candidate author labels alone. The magnitude of this improvement varies substantially across models, indicating that the effectiveness of sample-based definitions depends strongly on the underlying \acrshort{llm}.
Among the evaluated models, Qwen consistently achieves the strongest attribution performance across all three author subsets, reaching 43.3\% accuracy on the \cite{huang2024can} subset and 40.0\% on the randomly selected subset. Overall, Gemma generally outperforms Mixtral across the evaluated settings, demonstrating that the choice of the underlying \acrshort{llm} influences the effectiveness of sample-based definitions. Within each model, the choice of the \acrshort{aa} prompt also affects attribution performance. For a fixed author subset, the maximum observed difference between attribution prompts is 16.6 percentage points, whereas the maximum difference resulting from changing the underlying \acrshort{llm} reaches 36.6 percentage points. In addition, prompts achieving similar accuracy occasionally produce different weighted and macro F1 scores, indicating differences in class-wise attribution performance despite comparable overall accuracy. Although the LIP strategy proposed by \cite{huang2024can} performs competitively, it does not consistently achieve the highest performance across all evaluated models and author subsets. Instead, the best-performing attribution prompt varies depending on both the underlying \acrshort{llm} and the evaluated author subset. These findings indicate that, although both the underlying \acrshort{llm} and the attribution prompt influence performance, model selection has the greater overall impact on sample-based \acrshort{zs} \acrshort{aa}.

The choice of author subset also influences attribution performance. For both Mixtral and Gemma, the randomly selected subset yields the highest attribution accuracy, whereas the subset adopted from \cite{huang2024can} performs best only for Qwen. In contrast, the Top 10 author subset consistently produces lower performance across all three models. Interestingly, the subset adopted from \cite{huang2024can} does not consistently produce the highest attribution performance, suggesting that no single author subset is universally optimal for sample-based \acrshort{zs} \acrshort{aa}.These observations suggest that the effectiveness of sample-based definitions depends not only on the prompting strategy but also on the characteristics of the selected candidate authors and their representative writing samples.

Beyond attribution accuracy, the evaluated \acrshortpl{llm} exhibited noticeable differences in output validity and instruction-following behavior. Qwen consistently returned valid author identifiers for all evaluated samples, resulting in no `Unsure' predictions. Mixtral occasionally produced responses indicating that none of the provided candidate authors matched the query text or otherwise returned outputs that could not be mapped to a valid candidate identifier. In contrast, Gemma exhibited a different failure mode, frequently producing repetitive text or responses that did not follow the required output format, resulting in a larger number of `Unsure' predictions. Despite these instruction-following failures, Gemma generally outperformed Mixtral across the evaluated settings, indicating that output-format compliance and attribution accuracy are not necessarily coupled.

\begin{table}[!ht]
\centering
\small
\setlength{\tabcolsep}{2.5pt}
\renewcommand{\arraystretch}{1.05}

\caption{Definition-enhanced \acrshort{zs} \acrshort{aa} using representative writing samples across three sample-selection settings and three open-source \acrshortpl{llm}.}
\label{tab:zs_samples_csv_variants}

\resizebox{\columnwidth}{!}{%
\begin{tabular}{p{1.4cm} l c c c c c}
\toprule
\textbf{Subset} & \textbf{AA Prompt} & \textbf{Acc} & \textbf{W-F1} & \textbf{$\mu$F1} & \textbf{M-F1} & \textbf{Unsure} \\
\midrule

\multicolumn{7}{c}{\textbf{Mixtral}}\\
\midrule

\multirow{4}{=}{\cite{huang2024can} subset*}
& no\_guidance     & \textbf{16.6} & \textbf{11.9} & \textbf{16.6} & \textbf{11.0} & 8 \\
& little\_guidance &  6.6 &  4.1 &  6.6 &  3.7 & 11 \\
& grammar          & 10.0 &  6.5 & 10.0 &  5.8 & 10 \\
& LIP              & 13.3 &  9.7 & 13.3 &  8.8 & 6 \\
\midrule

\multirow{4}{=}{Top 10 authors}
& no\_guidance     & 10.0 & \textbf{7.5} & 10.0 & \textbf{7.0} & 1 \\
& little\_guidance & \textbf{13.3} & \textbf{7.5} & \textbf{13.3} & 6.9 & 3 \\
& grammar          & 10.0 & 4.3 & 10.0 & 4.1 & 1 \\
& LIP              & 10.0 & 5.6 & 10.0 & 5.6 & 0 \\
\midrule

\multirow{4}{=}{Random 10 authors}
& no\_guidance     & 13.3 &  9.6 & 13.3 &  9.1 & 8 \\
& little\_guidance & 13.3 &  9.8 & 13.3 &  8.6 & 9 \\
& grammar          & \textbf{23.3} & \textbf{17.3} & \textbf{23.3} & \textbf{15.7} & 4 \\
& LIP              & 16.6 & 12.1 & 16.6 & 11.7 & 4 \\

\midrule
\multicolumn{7}{c}{\textbf{Gemma}}\\
\midrule

\multirow{4}{=}{\cite{huang2024can} subset*}
& no\_guidance     &  6.6 &  6.6 &  6.6 &  6.0 & 1 \\
& little\_guidance & \textbf{16.6} & \textbf{13.3} & \textbf{16.6} & \textbf{12.2} & 15 \\
& grammar          & {16.6} & 10.7 & {16.6} & 9.8 & 9 \\
& LIP              & 10.0 & 10.0 & 10.0 & 9.0 & 19 \\
\midrule

\multirow{4}{=}{Top 10 authors}
& no\_guidance     & 13.3 &  8.6 & 13.3 &  8.6 & 0 \\
& little\_guidance & 13.3 & 10.8 & 13.3 & 10.8 & 0 \\
& grammar          & \textbf{16.6} & \textbf{14.1} & \textbf{16.6} & \textbf{14.1} & 0 \\
& LIP              & {16.6} & 10.9 & {16.6} & 10.9 & 0 \\
\midrule

\multirow{4}{=}{Random 10 authors}
& no\_guidance     & \textbf{26.6} & \textbf{23.3} & \textbf{26.6} & \textbf{23.3} & 8 \\
& little\_guidance & {26.6} & 22.2 & {26.6} & 21.0 & 8 \\
& grammar          & 23.3 & 17.4 & 23.3 & 16.8 & 9 \\
& LIP              & 10.0 & 7.2 & 10.0 & 7.2 & 12 \\

\midrule
\multicolumn{7}{c}{\textbf{Qwen}}\\
\midrule

\multirow{4}{=}{\cite{huang2024can} subset*}
& no\_guidance     & {43.3} & 35.5 & {43.3} & 35.5 & 0 \\
& little\_guidance & \textbf{43.3} & \textbf{36.1} & \textbf{43.3} & \textbf{36.1} & 0 \\
& grammar          & {43.3} & 33.8 &{43.3} & 33.8 & 0 \\
& LIP              & 40.0 & 33.8 & 40.0 & 33.8 & 0 \\
\midrule

\multirow{4}{=}{Top 10 authors}
& no\_guidance     & {23.3} & 15.8 & {23.3} & 15.80 & 0 \\
& little\_guidance & \textbf{23.3} & \textbf{16.1} & \textbf{23.3} & \textbf{16.1} & 0 \\
& grammar          & 16.6 & 11.3 & 16.6 & 11.3 & 0 \\
& LIP              & 13.3 & 8.3 & 13.3 & 8.3 & 0 \\
\midrule

\multirow{4}{=}{Random 10 authors}
& no\_guidance     & \textbf{40.0} & \textbf{30.4} & \textbf{40.0} & \textbf{30.4} & 0 \\
& little\_guidance & 33.3 & 26.3 & 33.3 & 26.3 & 0 \\
& grammar          & 30.0 & 23.5 & 30.0 & 22.8 & 0 \\
& LIP              & 30.0 & 21.2 & 30.0 & 21.2 & 0 \\

\bottomrule
\end{tabular}%
}
\footnotesize{* Mixtral results on \cite{huang2024can} subset differ slightly from those reported in the original paper.}
\end{table}

\begin{table}[!ht]
\centering
\small
\setlength{\tabcolsep}{4pt}
\renewcommand{\arraystretch}{1.08}
\caption{Definition-enhanced \acrshort{zs} \acrshort{aa} across three open-source \acrshortpl{llm}. Style descriptions are generated using 4 description prompts; P1(Baseline), P2(Content-agnostic), P3(Linguistic) and P4(Discriminative) explained in Table~\ref{tab:description_prompts} and evaluated using the attribution prompts in Table~\ref{tab:aa_prompts}.}
\label{tab:aa_desc_results}

\resizebox{\columnwidth}{!}{%
\begin{tabular}{llcccc}
\toprule
\textbf{Description P} & \textbf{AA Prompt} & \textbf{Acc (\%)} & \textbf{W-F1} & \textbf{$\mu$F1} & \textbf{M-F1} \\
\midrule

\multicolumn{6}{c}{\textbf{Mixtral}}\\
\midrule

\multirow{4}{*}{P1 }
& no\_guidance     & 23.3 & 14.3 & 23.3 & 14.3  \\
& little\_guidance & 23.3 & 14.7 & 23.3 & 14.7 \\
& grammar          & 23.3 & 15.2 & 23.3 & 15.2  \\
& LIP              & \textbf{26.6} & \textbf{16.0} & \textbf{26.6 }& \textbf{16.0}  \\
\midrule

\multirow{4}{*}{P2 }
& no\_guidance     & 16.6 &  6.7 & 16.6 &  6.7  \\
& little\_guidance & 20.0 & 10.3 & 20.0 &  9.5  \\
& grammar          & 20.0 &  9.2 & 20.0 &  9.2  \\
& LIP              & \textbf{20.0} & \textbf{10.3} & \textbf{20.0} & \textbf{10.3}  \\
\midrule

\multirow{4}{*}{P3 }
& no\_guidance     &  6.67 & 2.07 &  6.6 & 2.0 \\
& little\_guidance &  6.67 & 2.9 &  6.6 & 2.9  \\
& grammar          & 10.0 & 4.0 & 10.0 & 4.0  \\
& LIP              & \textbf{13.3} & \textbf{8.6} & \textbf{13.3} & \textbf{8.6}  \\
\midrule

\multirow{4}{*}{P4 }
& no\_guidance     & 23.3 & 15.9 & 23.3 & 15.1  \\
& little\_guidance & \textbf{26.6} & \textbf{17.2} & \textbf{26.67} & \textbf{17.25}  \\
& grammar          & 26.6 & 17.0 & 26.6 & 17.0  \\
& LIP              & 23.3 & 12.7 & 23.3 & 12.7  \\

\midrule
\multicolumn{6}{c}{\textbf{Gemma}}\\
\midrule

\multirow{4}{*}{P1 }
& no\_guidance     & 23.3 & 18.6 & 23.3 & 18.6  \\
& little\_guidance & \textbf{26.6} & \textbf{21.9} & \textbf{26.6} & \textbf{21.9}  \\
& grammar          & 23.3 & 16.9 & 23.3 & 16.9  \\
& LIP              & 23.3 & 14.8 & 23.3 & 14.8 \\
\midrule

\multirow{4}{*}{P2}
& no\_guidance     & 20.0 & 12.0 & 20.0 & 12.0  \\
& little\_guidance & \textbf{20.0} & \textbf{12.5} & \textbf{20.0} & \textbf{12.5}  \\
& grammar          & 20.0 & 11.4 & 20.0 & 11.4  \\
& LIP              & 20.0 & 11.7 & 20.0 & 11.7  \\
\midrule

\multirow{4}{*}{P3}
& no\_guidance     & 10.0 & 7.5 & 10.0 & 7.5  \\
& little\_guidance & 10.0 & 5.4 & 10.0 & 5.4  \\
& grammar          & \textbf{16.6} & \textbf{10.3} & \textbf{16.6} & \textbf{10.3 } \\
& LIP              & 10.0 & 5.1 & 10.0 & 5.1  \\
\midrule

\multirow{4}{*}{P4}
& no\_guidance     & 30.0 & 20.3 & 30.0 & 20.3  \\
& little\_guidance & 23.3 & 15.5 & 23.3 & 15.5  \\
& grammar          & 26.6 & 17.6 & 26.67 & 17.6  \\
& LIP              & \textbf{33.3} & \textbf{21.7} & \textbf{33.3} & \textbf{21.7}  \\

\midrule
\multicolumn{6}{c}{\textbf{Qwen}}\\
\midrule

\multirow{4}{*}{P1}
& no\_guidance     & 23.3 & 17.2 & 23.3 & 17.2  \\
& little\_guidance & 23.3 & 16.1 & 23.3 & 16.1  \\
& grammar          & 23.3 & 15.8 & 23.3 & 15.8  \\
& LIP              & \textbf{30.0} & \textbf{21.1} & \textbf{30.0} & \textbf{21.1}  \\
\midrule

\multirow{4}{*}{P2}
& no\_guidance     & \textbf{20.0} & \textbf{11.8} & \textbf{20.0} & \textbf{11.8}  \\
& little\_guidance & \textbf{20.0} & \textbf{11.8} & \textbf{20.0} & \textbf{11.8}  \\
& grammar          & 20.0 & 11.6 & 20.0 & 11.6  \\
& LIP              & 16.6 & 10.6 & 16.6 & 10.6  \\
\midrule

\multirow{4}{*}{P3 }
& no\_guidance     & \textbf{23.3} & \textbf{18.0} & \textbf{23.3} & \textbf{18.0}  \\
& little\_guidance & {13.3} & {8.8} & 13.3 & {8.8}  \\
& grammar          & {20.0} & {15.5} & 20.0 & {15.5}  \\
& LIP              & 13.3 & {10.2} & 13.3 & {10.2}  \\
\midrule

\multirow{4}{*}{P4}
& no\_guidance     & \textbf{33.3} & \textbf{22.8} & \textbf{33.3} & \textbf{22.8}  \\
& little\_guidance & \textbf{33.3} & \textbf{22.8} & \textbf{33.3} & \textbf{22.8}  \\
& grammar          & {33.3} & {23.3} & 33.3 & {23.3}  \\
& LIP              & 30.0 & 20.7 & 30.0 & 20.7  \\

\bottomrule
\end{tabular}
}
\end{table}

\subsection{Definition-Enhanced ZS AA: Descriptions as definition}

The third set of experiments replaces representative writing samples with LLM-generated descriptions of each author's writing style. Four different prompts were designed to generate these stylistic descriptions, which were subsequently used as author definitions for \acrshort{zs} \acrshort{aa}. Compared with representative writing samples, the generated descriptions provide a substantially more compact author representation, reducing the average representation size from 456.1 to 100 tokens per author, corresponding to a 78.1\% reduction in the number of tokens required to represent candidate authors. This experiment therefore investigates whether concise stylistic summaries generated using different description prompts retain sufficient information for effective \acrshort{zs} \acrshort{aa}. The corresponding attribution results are presented in Table~\ref{tab:aa_desc_results}.

Overall, description-based definitions achieve lower attribution performance than representative writing samples. While the highest attribution accuracy obtained using representative writing samples reaches 43.3\%, the best description-based configuration achieves 33.3\%. This reduction in performance is accompanied by a substantial decrease in representation size, suggesting a trade-off between prompt efficiency and attribution accuracy. Although concise style descriptions capture important stylistic characteristics, they do not preserve the same level of discriminative information contained in representative writing samples. One possible explanation is that each style description was generated from a single representative writing sample to ensure a fair comparison across all definition strategies. Generating descriptions from multiple representative samples may capture a broader range of author-specific stylistic characteristics, potentially leading to richer author representations and improved attribution performance, although this remains to be investigated.

Across the evaluated description prompts, clear differences are observed in attribution performance. Prompt P4 (Discriminative) consistently achieves the strongest or jointly strongest performance across all three evaluated \acrshort{llm}s, reaching the highest accuracy for Mixtral (26.6\%), Gemma (33.3\%), and Qwen (33.3\%). Prompt P1 (Baseline) also performs consistently well, frequently achieving performance comparable to P4 across all three evaluated LLMs. Interestingly, this suggests that relatively simple stylistic descriptions are already sufficient to capture useful author-specific information for downstream \acrshort{aa}. In contrast, Prompt P3 (Linguistic) generally produces the weakest attribution performance, particularly for Mixtral and Gemma, while Prompt P2 (Content-agnostic) exhibits relatively stable but lower performance. These observations suggest that the formulation of the description-generation prompt substantially influences the quality of the generated author representations and, consequently, downstream attribution performance. 

The underlying \acrshort{llm} continues to influence attribution performance, although the performance gap between models becomes smaller than in the sample-based setting. While Qwen remains among the best-performing models, Gemma achieves comparable attribution performance under several prompt configurations, indicating that concise style descriptions reduce, but do not eliminate, differences between the evaluated \acrshort{llm}s. Within each description prompt, the choice of the \acrshort{aa} prompt also influences attribution performance, although no single attribution prompt consistently achieves the highest performance across all description prompts and models.

Unlike the sample-based experiments, all evaluated models consistently returned valid author identifiers in the description-based setting, resulting in no ``Unsure'' predictions. This suggests that the use of concise style descriptions is associated with improved output-format compliance while simultaneously reducing the amount of author-specific context required for attribution.

\subsection{Definition-Enhanced ZS AA: Style embeddings as definition}

Our final experiments directly use style embeddings as definitions. We first evaluate \acrshort{aa} using LISA embeddings in a single-stage setting. 
Although the single-stage approach achieves 36.6\% accuracy, the Top-$k$ recall results indicate that the correct author is frequently retained among the highest-ranked candidates. Specifically, Top-2, Top-3 and Top-4 recall reach 56.6\%, 73.3\% and 76.6\%, respectively. These observations suggest that the embedding representation often places the true author close to the top of the candidate ranking, even when it is not selected as the top prediction. Motivated by this behavior, and inspired by the use of a two-stage framework for \acrshort{aa} in prior work \cite{tan2025open}, we investigate a two-stage approach, described in Section~3.4, in which the first stage reduces the candidate author space based on cosine similarity, while the second stage refines attribution using selected subsets of the style embedding dimensions.

Table~\ref{tab:lisa_blog_combined_results} summarizes the performance of LISA-based \acrshort{aa} under single-stage and two-stage settings using different values of $k$ and $N$. The single-stage baseline achieves 36.6\% accuracy. To investigate the effect of candidate-space reduction, we first fix $N=20$ and vary $k$. Under this setting, the two-stage framework consistently improves attribution performance over the single-stage baseline, with the highest accuracy of 43.3\% achieved using presence-based feature selection ($P$) at $k=2$. Although Stage-1 recall increases from 56.6\% to 76.6\% as $k$ increases from 2 to 4, the final attribution accuracy does not improve accordingly, suggesting that enlarging the candidate set beyond the highest-ranked candidates provides limited additional benefit.

To evaluate the effect of feature selection, we next fix $k=4$ and vary $N$. Attribution accuracy improves as $N$ increases from 20 to 35, with the best result of 50.0\% achieved using the combined presence-and-absence strategy (P+A). However, performance decreases when $N$ is further increased to 45, suggesting that incorporating additional embedding dimensions eventually introduces less informative features that reduce discriminative power. Based on  experiments which identified $k=2$ as the most effective candidate-space size, we performed an additional experiment combining $k=2$ with the best-performing feature-selection setting ($N=35$). This additional experiment further  increased attribution accuracy to 56.6\%  using the combined presence-and-absence strategy (P+A), indicating that jointly optimizing candidate-space reduction and embedding-dimension selection yields the strongest overall performance. Overall, these findings indicate that jointly optimizing candidate-space reduction and embedding-dimension selection substantially improves attribution performance, while the effectiveness of the second stage remains dependent on the feature-selection strategy employed.

Overall, among all evaluated definition strategies, style embeddings achieve the highest attribution performance, reaching 56.6\% accuracy and outperforming LLMs with representative writing samples (43.3\%) and LLM-generated style descriptions (33.3\%).

\begin{table}[!ht]
\centering
\small
\setlength{\tabcolsep}{2.5pt}
\renewcommand{\arraystretch}{1.03}

\caption{Definition-Enhanced ZS AA: Style embeddings as definitions under different values of $k$ and $N$. P denotes presence and A denotes absence-based selection.}
\label{tab:lisa_blog_combined_results}

\resizebox{\columnwidth}{!}{%
\begin{tabular}{c l c c c c c}
\toprule
\textbf{Setting} &
\textbf{Method} &
\shortstack{\textbf{Stage-1}\\\textbf{Recall (\%)}} &
\textbf{Acc (\%)} &
\textbf{W-F1} &
\textbf{$\mu$F1} &
\textbf{M-F1} \\
\midrule

\multirow{4}{*}{$k$=2, $N$=20}
& Single-stage & -- & 36.6 & 34.67 & 36.6 & 34.6 \\
& 2-stage: P   & 56.67 & \textbf{43.3} & \textbf{44.5} & \textbf{43.3} & \textbf{44.5} \\
& 2-stage: A   & 56.6 & 40.0 & 37.57 & 40.0 & 37.5\\
& 2-stage: P+A & 56.6 & 36.6 & 35.1 & 36.6 & 35.1 \\
\midrule
\multirow{4}{*}{$k$=3, $N$=20}
& Single-stage & -- & 36.6 & 34.6 & 36.6 & 34.6 \\
& 2-stage: P   & \textbf{73.3} & \textbf{40.0} & \textbf{39.2} & \textbf{40.0} & \textbf{39.2} \\
& 2-stage: A   & 73.3 & 36.6 & 34.3 & 36.6 & 34.3 \\
& 2-stage: P+A & 73.3 & 33.3 & 31.9 & 33.3 & 31.9 \\
\midrule

\multirow{4}{*}{$k$=4, $N$=20}
& Single-stage & -- & 36.6 & 34.6 & 36.6 & 34.6 \\
& 2-stage: P   & \textbf{76.67} & \textbf{40.0} & \textbf{39.31} & \textbf{40.0} & \textbf{39.3} \\
& 2-stage: A   & 76.6 & 33.3 & 32.2 & 33.3 & 32.2 \\
& 2-stage: P+A & 76.6 & 30.00 & 29.3 & 30.0 & 29.3 \\
\midrule

\multirow{4}{*}{$k$=4, $N$=35}
& Single-stage & -- & 36.6 & 34.6 & 36.6 & 34.6 \\
& 2-stage: P   & 76.6 & 40.0 & 36.0 & 40.0 & 36.0 \\
& 2-stage: A   & 76.6 & 46.6 & 43.7 & 46.6 & 43.7 \\
& 2-stage: P+A & \textbf{76.6} & \textbf{50.0} & \textbf{46.38} & \textbf{50.0} & \textbf{46.3} \\
\midrule

\multirow{4}{*}{$k$=4, $N$=45}
& Single-stage & -- & \textbf{36.6} & \textbf{34.6} & \textbf{34.6} & \textbf{34.6} \\
& 2-stage: P   & 76.6 & 36.6 & 32.7 & 36.6 & 32.7 \\
& 2-stage: A   & 76.6 & 33.3 & 28.9 & 33.3 & 28.9 \\
& 2-stage: P+A & \textbf{76.6} & {33.3} & {28.9} & {33.3} & {28.9} \\
\midrule

\multirow{4}{*}{$k$=2, $N$=35}
& Single-stage & -- & 36.6 & 34.6 & 36.6 & 34.6 \\
& 2-stage: P   & 56.6 & 43.3 & 44.5 & 43.3 & 44.5 \\
& 2-stage: A   & 56.6 & 40.0 & 37.5 & 40.0 & 37.5 \\
& 2-stage: P+A & \textbf{56.6} & \textbf{56.6} & \textbf{57.0} & \textbf{56.6} & \textbf{57.0} \\
\bottomrule
\end{tabular}%
}
\end{table}

\subsection{Comparative analysis}
\begin{figure}[t]
    \centering
    \includegraphics[width=\columnwidth]{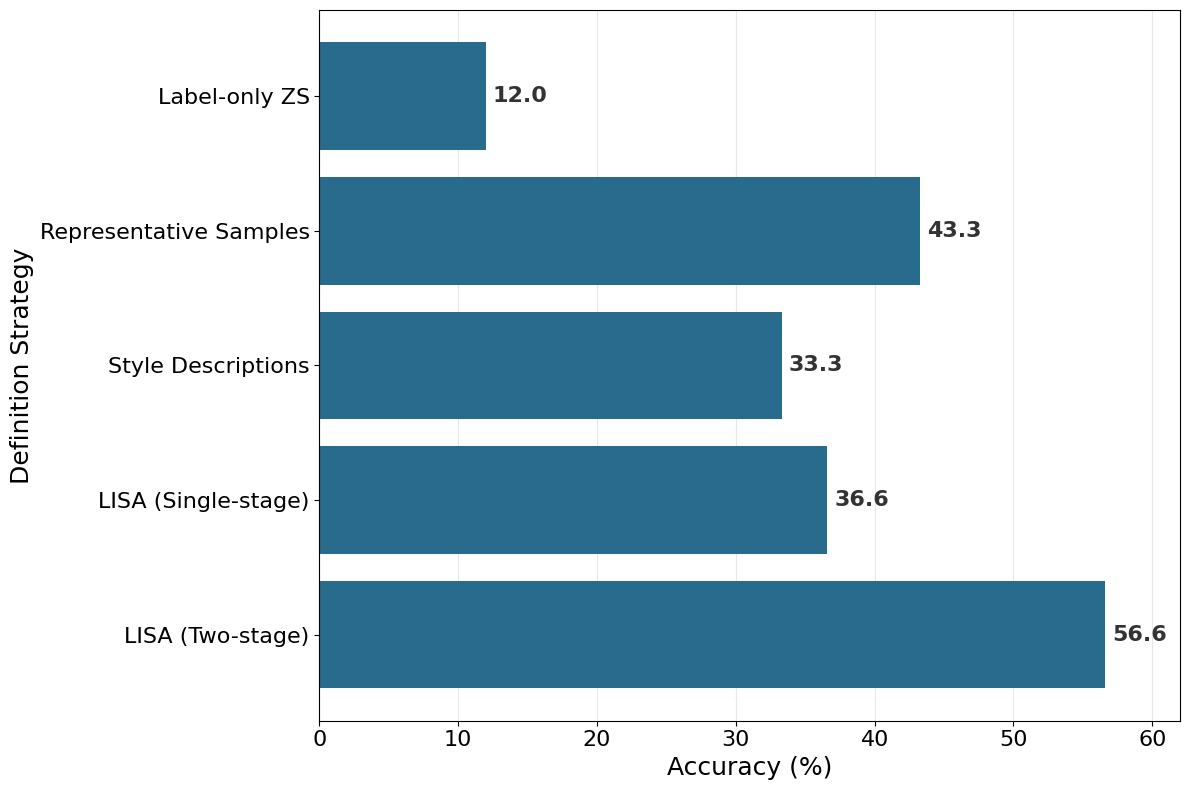}
    \caption{Comparison of the highest attribution accuracy achieved across the evaluated \acrshort{aa} approaches.}
    \label{fig:definition_summary}
\end{figure}

Figure~\ref{fig:definition_summary} compares the highest attribution accuracy achieved by each evaluated \acrshort{zs} \acrshort{aa} approach. In the label-only \acrshort{zs} setting, attribution performance remains close to chance-level performance, indicating that candidate author labels alone provide insufficient information for reliable stylistic attribution. Introducing author-specific representations consistently improves performance over the label-only baseline, demonstrating the importance of providing auxiliary author-specific information. Representative writing samples substantially improve attribution accuracy but remain sensitive to the selected author subsets and representative examples. LLM-generated style descriptions provide a considerably more compact author representation while sacrificing some attribution performance relative to representative writing samples. Leaving LLMs and using style embeddings with cosine similarity substantially improve attribution performance over the preceding approaches. While the single-stage LISA framework already outperforms the label-only baseline and style descriptions, the proposed two-stage refinement further improves attribution performance, achieving the strongest overall performance among the evaluated approaches. 

\section{Conclusion}

This paper investigated \acrshort{zs} \acrshort{aa} using different author representation strategies, including label-only prompting, representative writing samples, LLM-generated  descriptions, and style embeddings. Beyond evaluating individual approaches, this work provides a unified comparison of how different author representations influence \acrshort{zs} \acrshort{aa}. The results show that label-only \acrshort{zs} \acrshort{aa} remains ineffective, whereas incorporating author-specific representations consistently improves attribution performance without task-specific fine-tuning.

Among the evaluated approaches, style embeddings proved to be the most effective representation, with the proposed two-stage LISA framework further improving attribution performance over the single-stage baseline through candidate-space reduction and embedding-dimension selection. In contrast, the LLM-based approaches achieved lower attribution performance, indicating that \acrshort{zs} \acrshort{aa} with current open-source LLMs remains challenging.

Overall, this work highlights author representation as a fundamental challenge in \acrshort{zs} \acrshort{aa} and demonstrates the potential of embedding-based representations for addressing it. Future work should investigate richer stylistic representations, transfer learning and parameter-efficient fine-tuning, improved candidate retrieval and embedding-dimension selection strategies, and evaluation across larger and more diverse \acrshort{aa} datasets.

\section*{Acknowledgments}
This work is partly supported by the Wallenberg AI, Autonomous Systems and Software Program (WASP), funded by Knut and Alice Wallenberg Foundations and counterpart funding from Lulea University of Technology (LTU).


\FloatBarrier
\appendix

\section{Limitations and Future Work}

Although this study systematically compares multiple \acrshort{zs} \acrshort{aa} approaches across three open-source \acrshort{llm}s, the evaluation remains limited in scope. The experiments were conducted on a blog \acrshort{aa} dataset, and the observed findings may not generalize to other writing genres, languages, or \acrshort{aa} benchmarks. Future work should therefore evaluate the proposed approaches across more diverse datasets with varying stylistic characteristics and numbers of candidate authors.

For a fair comparison across the evaluated author representation strategies, both the LLM-generated style descriptions and the LISA style embeddings were derived from a single representative writing sample for each author. While this design isolates the effect of the representation itself, relying on a single sample may not fully capture the stylistic diversity of an author's writing. Future work should investigate representations generated from multiple representative writing samples, which may produce richer and more representative author representations and further improve attribution performance.

The proposed two-stage embedding framework employs relatively simple candidate refinement and embedding-dimension selection strategies based on presence and absence information. Although this approach substantially improves attribution performance over the single-stage baseline, more principled feature-selection methods, learned similarity functions, or adaptive embedding refinement strategies may further improve attribution performance.

Overall, this work demonstrates that the choice of author representation is a key factor in \acrshort{zs} \acrshort{aa}. Future research should therefore focus on developing richer and more transferable stylistic representations while exploring more effective embedding-based retrieval and refinement strategies across a broader range of \acrshort{llm}s and \acrshort{aa} datasets.

\section{Prompts Used in this Study}
\label{app:prompts}
For ZS AA without definition, the prompts used are shown in Table \ref{tab:zs_label_only_prompt}. For description generation we created 4 prompts, and they are  shown in Table \ref{tab:description_prompts}. Finally for AA the prompts that are used are proposed by \cite{huang2024can} and are shown in Table \ref{tab:aa_prompts}.

\label{sec:prompts}
\begin{table}[t]
\centering
\small
\caption{\acrshort{zs}-shot (ZS) label-only prompt without definitions used for \acrshort{aa}.}
\label{tab:zs_label_only_prompt}

\begin{tabular}{p{0.22\linewidth} p{0.72\linewidth}}
\hline
\textbf{Prompt Type} & \textbf{Description} \\
\hline

\textbf{ZS (Label-only)} &
You are a helpful assistant for author attribution. Given the following text, identify who is the most likely author. Choose ONLY from the following IDs: \textless author IDs\textgreater. Text: \textless text\textgreater. Answer: \\

\hline

\end{tabular}
\end{table}
\begin{table*}[t]
\centering
\small
\caption{Prompts used for generating author descriptions.}
\begin{tabular}{p{0.12\linewidth} p{0.82\linewidth}}
\hline
\textbf{Prompt} & \textbf{Description} \\
\hline

P1 &
you are an expert in writing style extraction, Given the writing samples of the author, extract writing style from it and write it in one consise paragraph. \\

\hline

P2 &
P1 + Focus purely on \textit{how} the author writes, not \textit{what} they write. \\

\hline

P3 &
P1 + Pay special attention to stylistic signals such as sentence length, grammatical looseness or rigidity, punctuation patterns, slang usage, formality level, and lexical choices. \\

\hline

P4 &
P1 + Highlight the most distinctive stylistic features that uniquely identify the author and distinguish them from others. \\

\hline

\end{tabular}

\label{tab:description_prompts}
\end{table*}

\begin{table*}[t]
\centering
\small
\caption{AA prompts adopted from \cite{huang2024can}.}
\label{tab:aa_prompts}

\begin{tabular}{p{0.18\linewidth} p{0.76\linewidth}}
\hline
\textbf{Prompt} & \textbf{Description} \\
\hline

\textbf{System Instruction} &
Respond with a JSON object including two elements: ``analysis'' (reasoning behind the answer) and ``answer'' (the predicted author ID of the query text). \\

\hline

\textbf{No Guidance} &
Given a set of texts with known authors and a query text, determine the author of the query text. Input query text: \textless query text\textgreater; Texts from potential authors: \textless candidate texts\textgreater. \\

\hline

\textbf{Style Guidance} &
Given a set of texts with known authors and a query text, determine the author of the query text. Do not consider topic differences. Input query text: \textless query text\textgreater; Texts from potential authors: \textless candidate texts\textgreater. \\

\hline

\textbf{Grammar Guidance} &
Given a set of texts with known authors and a query text, determine the author of the query text. Focus on grammatical styles. Input query text: \textless query text\textgreater; Texts from potential authors: \textless candidate texts\textgreater. \\

\hline

\textbf{LIP} &
Given a set of texts with known authors and a query text, determine the author of the query text. Analyze writing styles while disregarding topic and content. Focus on linguistic features such as phrasal verbs, modal verbs, punctuation, rare words, affixes, quantities, humor, sarcasm, typographical errors, and misspellings. Input query text: \textless query text\textgreater; Texts from potential authors: \textless candidate texts\textgreater. \\

\hline

\end{tabular}
\end{table*}

\textbf{Description Prompts}:

\textbf{P1: Baseline prompt.} This baseline prompt instructs the model to extract the author's writing style from the provided representative writing sample and summarize it in a concise paragraph. It serves as the reference prompt against which the remaining prompt variants are compared. 

\textbf{P2: Content-agnostic style descriptions.} This prompt explicitly instructs the model to focus on \emph{how} the author writes rather than \emph{what} they write. By discouraging content-based descriptions, it encourages more topic-independent stylistic representations.

\textbf{P3: Linguistically focused style descriptions.} This prompt explicitly directs the model to consider linguistic characteristics such as sentence structure, punctuation, lexical choice, slang usage, and formality. Specifying these stylistic dimensions encourages more structured and feature-oriented author representations.

\textbf{P4: Discriminative stylistic fingerprint.} This prompt encourages the model to identify the author's most distinctive stylistic characteristics by emphasizing features that uniquely distinguish the author from others. Rather than producing a general stylistic summary, it aims to generate a more discriminative stylistic fingerprint.

\bibliographystyle{apalike}
{\small
\bibliography{example}}
\end{document}